\documentclass{article} 
\usepackage{hyperref}
\usepackage{graphicx}
\usepackage{geometry}
\usepackage{epstopdf}
\usepackage{amssymb}
\usepackage{amsmath}

\graphicspath{{./}}

\title{Critical Parameters in Particle Swarm Optimisation}

\author{ J.~Michael Herrmann\footnote{corresponding author: j.michael.herrmann@gmail.com}, Adam Erskine, Thomas Joyce\\
Institute for Perception, Action and Behaviour\\
School of Informatics, The University of Edinburgh\\
10 Crichton St, Edinburgh EH8 9AB, Scotland, U.K.}

\date{}

\begin{document}

\maketitle

\begin{abstract}
	Particle swarm optimisation is a metaheuristic algorithm
	which finds reasonable solutions in a wide range of applied problems
	if suitable parameters are used. We study the properties of the
	algorithm in the framework of random dynamical systems which, due to 
	the quasi-linear swarm dynamics, yields analytical results for 
	the stability properties of the particles. 
	Such considerations predict a relationship between the parameters of the algorithm
	that marks the edge between convergent and divergent behaviours. 
	Comparison with simulations 
	indicates that the algorithm performs best near this margin of instability. 
\end{abstract}

\section{PSO Introduction}

Particle Swarm Optimisation (PSO,~\cite{kennedy1995particle}) is a metaheuristic algorithm 
which is widely used to solve search and optimisation tasks. 
It employs a number of particles as a swarm of potential solutions.
Each particles shares knowledge about the current overall best solution and also retains a memory of 
the best solution it has encountered itself previously. Otherwise the particles, after random initialisation,
obey a linear dynamics of the following form
\begin{eqnarray}
\label{eq:PSOVelocityUpdate}
\mathbf{v}_{i,t+1}&=&\omega\mathbf{v}_{i,t}+\alpha_{2} \mathbf{R}_{1}(\mathbf{p}_{i}-\mathbf{x}_{i,t})+\alpha_{2} \mathbf{R}_{2}(\mathbf{g}-\mathbf{x}_{i,t})
\cr
\mathbf{x}_{i,t+1}&=&\mathbf{x}_{i,t}+\mathbf{v}_{i,t+1}
\label{eq:PSOPositionUpdate}
\end{eqnarray}
Here $\mathbf{x}_{i,t}$ and $\mathbf{v}_{i,t}$, $i=1,\dots,N$, $t=0,1,2,\dots$,
represent, respectively, the $d$-dimensional position in the search space and the velocity 
vector of the $i$-th particle in the swarm at time $t$. 
The velocity update
contains an inertial term parameterised by $\omega$ and includes attractive forces 
towards the personal best location $\mathbf{p}_{i}$  and towards 
the globally
best location $\mathbf{g}$, which are parameterised by $\alpha_{1}$ and
and $\alpha_{2}$, respectively. 
The symbols $\mathbf{R}_{1}$ and $\mathbf{R}_{2}$ denote diagonal matrices whose non-zero entries 
are uniformly distributed in the unit interval. The number of particles $N$ is quite low in 
most applications, usually amounting to a few dozens.

In order to function as an optimiser, the algorithm uses a nonnegative cost function 
$F:\mathbb{R}^d \to \mathbb{R}$, where without loss of generality $F(\mathbf{x}^*)=0$ 
is assumed at an optimal solution $\mathbf{x}^*$. In many problems, where PSO is applied, there are also states with near-zero costs can be considered as good solutions.
The cost function is evaluated for the state of each particle at each time step. 
If $F(\mathbf{x}_{i,t})$ is better than $F(\mathbf{p}_{i})$, then the personal
best $\mathbf{p}_{i}$ is replaced by $\mathbf{x}_{i,t}$. 
Similarly, if one of the particles arrives at a state
with a cost less than $F(\mathbf{g})$, then $\mathbf{g}$ is replaced in all particles by the 
position of the particle that has discovered the new solution. 
If its velocity is non-zero, a particle will depart from the current best location,
but it may still have a chance to return guided by the force terms in the dynamics.

Numerous modifications and variants have been 
proposed since the algorithm's inception \cite{kennedy1995particle} and it continues to enjoy 
widespread usage. Ref.~\cite{poli2008analysis} groups around 700 PSO papers into 26 discernible
application areas. Google Scholar reveals over 150,000 results for ``Particle Swarm Optimisation''
in total and 24,000 for the year 2014.

In the next section we will report observations from a simulation of a particle swarm and move on to
a standard matrix formulation of the swarm dynamics in order to describe some of the existing analytical 
work on PSO. In Sect.~\ref{Critical} we will argue for a formulation of PSO as a random dynamical system
which will enable us to derive a novel exact characterisation of the dynamics of one-particle system, 
which will then be generalised towards the more realistic case of a multi-particle swarm.
In Sect.~\ref{Simulations} we will compare the theoretical predictions with simulations on
a representative set of benchmark functions. Finally, in Sect.~\ref{discussion} we will discuss the
assumption we have made in the theoretical solution in Sect.~\ref{Critical} and address the applicability
of our results to other metaheuristic algorithms and to practical optimisation problems.

\section{Swarm dynamics}

\subsection{Empirical properties}

The success of the algorithm in locating good solutions depends on the dynamics of the particles 
in the state space of the problem. In contrast to many evolution strategies, it is not straight 
forward to interpret the particle swarm as following a landscape defined by the cost function. 
Unless the current best
positions $\mathbf{p}$ or $\mathbf{g}$ change, the particles do not interact with each other and 
follow an intrinsic dynamics that does not even indirectly obtain any gradient information.

The particle dynamics depends on the parameterisation of the Eq.~\ref{eq:PSOVelocityUpdate}.
To obtain the best result one needs to select parameter settings that 
achieve a balance between the particles exploiting the knowledge of good known locations and exploring regions of the problem space that have not been visited before. Parameter values often need to be experimentally determined, and poor selection may result in premature convergence of the swarm to poor local minima or in a divergence of the particles towards regions that are irrelevant for the problem.

\begin{figure}[th]
\noindent \begin{centering}
\includegraphics[width=0.75\linewidth]{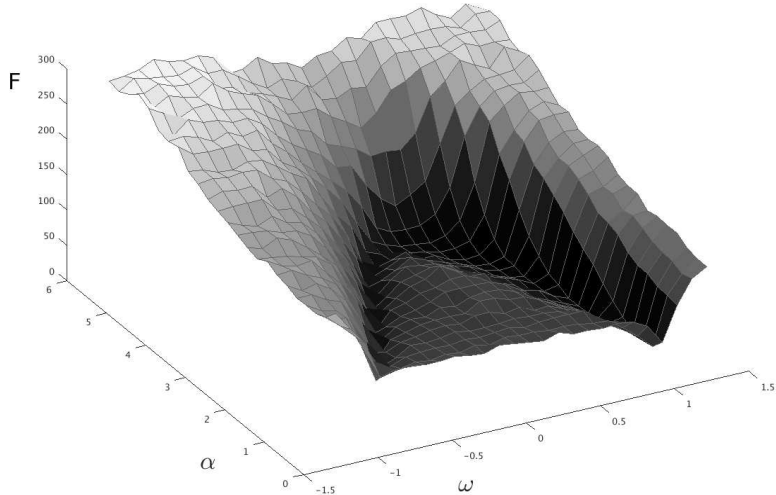}
\par\end{centering}
\vspace{-5mm}
\caption{Typical PSO performance as a function of its $\omega$ and $\alpha$ parameters. 
Here a 25 particle swarm was run for pairs of $\omega$ and $\alpha$ values ($\alpha_1=\alpha_2=\alpha/2$). 
Cost function here was the $d=10$ non-continuous rotated Rastrigin function~\cite{CEC2013}. 
Each parameter pair was repeated 25 times and the minimal costs after 2000 iterations were 
averaged. \label{fig:PSOPerformance}}
\end{figure}

Empirically we can execute PSO against a variety of problem functions with a range of $\omega$ and 
$\alpha_{1,2}$ values. Typically the algorithm  shows performance of the form depicted 
in Fig.~\ref{fig:PSOPerformance}. The best solutions found show a curved relationship 
between $\omega$ and $\alpha=\alpha_1+\alpha_2$, with $\omega\approx 1$ at small $\alpha$, and
$\alpha\gtrapprox 4$ at small $\omega$. 
Large values of both $\alpha$ and $\omega$ are found 
to cause the particles to diverge leading to results far from optimality, while at small values for both
parameters the particles converge to a nearby solution which sometimes is acceptable. 
For other cost functions similar relationships are observed in numerical tests (see Sect.~\ref{Simulations})
unless no good solutions found due to problem complexity or run time limits, see Sect.~\ref{runtime}.
For simple cost functions, such as a single well potential, there are also parameter combinations with 
small $\omega$ and small $\alpha$ will usually lead to good results.
The choice of $\alpha_1$ and $\alpha_2$ at constant $\alpha$ may
have an effect for some cost functions, but does not seem to have a big effect in most cases.

\subsection{Matrix formulation}

In order to analyse the behaviour of the algorithm it is convenient to use a matrix formulation 
by inserting the velocity explicitly in the second equation (\ref{eq:PSOPositionUpdate}).
\begin{equation}
\mathbf{z}_{t+1}=M \mathbf{z}_t + \alpha_1 \mathbf{R}_1 (\mathbf{p},\mathbf{p})^{\top}+
\alpha_2 \mathbf{R}_2 (\mathbf{g},\mathbf{g})^{\top} \label{eq:matrix_from}
\end{equation}
with $\mathbf{z}=(\mathbf{v},\mathbf{x})^{\top}$ and 
\begin{equation}
M=\left(\begin{array}{cc} 
			\omega \mathbf{I}_d & -\alpha_1 \mathbf{R}_1- \alpha_2 \mathbf{R}_2 \\
			\omega \mathbf{I}_d & \mathbf{I}_d-\alpha_1 \mathbf{R}_1- \alpha_2 \mathbf{R}_2 
	\end{array}\right),
\label{matrix}
\end{equation}
where $\mathbf{I}_d$ is the unit matrix in $d$ dimensions. Note that the two occurrence of $\mathbf{R}_1$ in 
Eq.~\ref{matrix} refer to the same realisation of the random variable. Similarly, the two $\mathbf{R}_2$'s are 
the same realisation, but different from $\mathbf{R}_1$.
Since the second and third term on the right in Eq.~\ref{eq:matrix_from} are 
constant most of the time, the analysis of the algorithm can focus on the properties of the matrix $M$.
In spite of its wide applicability, PSO has not been subject to deeper theoretical study, which may
be due to the multiplicative noise in the simple quasi-linear, quasi-decoupled dynamics. In previous
studies the effect of the noise has largely been ignored.

\subsection{Analytical results}

An early exploration of the PSO dynamics~\cite{kennedy1998behavior} considered a single particle in 
a one-dimension space where the personal and global best locations were taken to be the same. 
The random components were replaced by their averages such that apart from random initialisation
the algorithm was deterministic. 
Varying the parameters was shown to result 
in a range of periodic motions and divergent behaviour for the case of $\alpha_1+\alpha_2\ge4$. 
The addition of the random vectors was seen as beneficial as it adds noise to the deterministic search.

Control of velocity, not requiring the enforcement of an arbitrary maximum value as in 
Ref.~\cite{kennedy1998behavior}, is derived in an analytical manner by~\cite{clerc2002particle}. 
Here eigenvalues derived from the dynamic matrix of a simplified version of the PSO algorithm 
are used to imply various search behaviours. Thus, again the  $\alpha_1+\alpha_2\ge4$ case is
expected to diverge. For $\alpha_1+\alpha_2<4$ various cyclic and quasi-cyclic motions are shown 
to exist for a non-random version of the algorithm.

In Ref.~\cite{trelea2003particle} again a single particle was considered in a one dimensional 
problem space, using a deterministic version of PSO, setting $\mathbf{R}_{1}=\mathbf{R}_{2}=0.5$. 
The eigenvalues of the system were determined as functions of $\omega$ and a combined $\alpha$, which
leads to three conditions: The particle is shown to converge when $\omega<1$, $\alpha>0$ and 
$2\omega-\alpha+2>0$. Harmonic oscillations occur for 
$\omega^2+\alpha^2-2\omega\alpha-2\omega-2\alpha+1<0$ 
and a zigzag motion is expected if 
$\omega<0$ and $\omega-\alpha+1<0$. 
As with the preceding papers the discussion of the random numbers in the algorithm views 
them purely as enhancing the search capabilities by adding a \emph{drunken walk} 
to the particle motions. Their replacement by expectation values was thus believed to 
simplify the analysis with no loss of generality. 

We show in this contribution that the 
iterated use of these random factors $\mathbf{R}_{1}$ and $\mathbf{R}_{2}$
in fact adds a further level of complexity to the 
dynamics of the swarm which affects the behaviour of the algorithm in a non-trivial way. In Ref.~\cite{jiang2007stagnation} these factors were given some consideration. Regions of convergence and divergence separated by a curved line were predicted. This line separating these regions (an equation for which is given in Ref.~\cite{cleghorn2014generalized}) fails to include some parameter settings that lead to convergent swarms. Our analytical solution of the stability problem for the swarm dynamics explains
why parameter settings derived from the deterministic approaches are not in line with
experiences from practical tests. For this purpose we will now formulate the PSO algorithm as
a random dynamical system and present an analytical solution for the swarm dynamics in a
simplified but representative case.

\section{Critical swarm conditions for a single particle\label{Critical}}

\subsection{PSO as a random dynamical system}

As in Refs.~\cite{kennedy1998behavior,trelea2003particle} the dynamics of the particle swarm will 
be studied here as well in the single-particle case. This can be justified
because the particles interact only
via the global best position such that, while $\mathbf{g}$ (\ref{eq:PSOVelocityUpdate}) is unchanged,
single particles exhibit qualitatively the same dynamics as in the swarm. 
For the one-particle case we have necessarily $\mathbf{p}=\mathbf{g}$,
such that shift invariance allows us to set both to zero, which leads us to the following
is given by the stochastic-map  formulation of the PSO dynamics (\ref{eq:matrix_from}).
\begin{equation}
\mathbf{z}_{t+1}=M \mathbf{z}_t 
	\label{eq:pso_expl-1}
\end{equation}
Extending earlier approaches we will explicitly consider the randomness of the dynamics,
i.e.~instead of averages over $\mathbf{R}_1$ and $\mathbf{R}_2$ we consider a random
dynamical system with dynamical matrices $M$ chosen from the set
\begin{equation}
	{\cal M}_{\alpha,\omega}=\left\{
\left(\begin{array}{cc} 
			\omega \mathbf{I}_d & -\alpha \mathbf{R} \\
			\omega \mathbf{I}_d & \mathbf{I}_d-\alpha \mathbf{R} 
	\end{array}\right)\!,\,\,
	\mathbf{R}_{ij}=0 \mbox{ for } i\ne j  \mbox{ and } \mathbf{R}_{ii}\in\left[0,1\right], 
\right\} 
\label{eq:matrix_set}
\end{equation}
with $\mathbf{R}$ being in both rows the same realisation of a random diagonal matrix 
that combines the effects of $\mathbf{R}_1$ and $\mathbf{R}_2$ (\ref{eq:PSOPositionUpdate}).
The parameter $\alpha$ is the sum $\alpha_1+\alpha_2$ with $\alpha_1,\alpha_2\ge0$ and $\alpha >0$.
As the diagonal elements of $\mathbf{R}_1$ and $\mathbf{R}_2$ are 
uniformly distributed in $\left[0,1\right]$, the distribution of the random variable 
$\mathbf{R}_{ii} = \frac{\alpha_1}{\alpha} \mathbf{R}_{1,ii} + \frac{\alpha_2}{\alpha} \mathbf{R}_{2,ii}$ 
in Eq.~\ref{eq:pso_expl-1} is given by a convolution
of two uniform random variables, namely
\begin{equation}
P_{\alpha_1,\alpha_2}(r)=\begin{cases}
	\frac{\alpha r}{\max\{\alpha_1,\alpha_2\}}&\mbox{if } 0\le r\le \min\{\frac{\alpha_1}{\alpha},\frac{\alpha_2}{\alpha}\}\\
	\frac{\alpha}{\max\{\alpha_1,\alpha_2\}} & \mbox{if } \min\{\frac{\alpha_1}{\alpha},\frac{\alpha_2}{\alpha}\}<r\le\max\{\frac{\alpha_1}{\alpha},\frac{\alpha_2}{\alpha}\}\\
	\frac{\alpha \left(\alpha-r\right)}{\alpha_1 \alpha_2} & \mbox{if } \max\left\{\frac{\alpha_1}{\alpha},\frac{\alpha_2}{\alpha}\right\}<r\le 1
\end{cases}
\label{eq:convolution}
\end{equation}
if the variable $r\in [0,1]$ and $P_{\alpha_1,\alpha_2}(r)=0$ otherwise.
$P_{\alpha_1,\alpha_2}(r)$ has a tent shape for $\alpha_1=\alpha_2$ and a box shape in the 
limits of either $\alpha_1\to 0$ or $\alpha_2\to 0$. 
The case $\alpha_1=\alpha_2=0$, where the swarm does not obtain information about the fitness function,
will not be considered here.

We expect that the multi-particle
PSO is well represented by the simplified version for $\alpha_2\gg\alpha_1$ or $\alpha_1\gg\alpha_2$,
the latter case being irrelevant in practice. For $\alpha_1\approx \alpha_2$ deviations from the
theory may occur because in the multi-particle case $\mathbf{p}$ and $\mathbf{g}$ will be different for most particles. We will discuss this as well as the effects of the switching of the
dynamics at discovery of better solutions in Sect.~\ref{sub:The-role-of}.

\subsection{Marginal stability\label{sub:Stationary-distribution}}

While the swarm does not discover any new solutions, its dynamical properties are 
determined by an infinite product of matrices from the set ${\cal M}$ (\ref{eq:matrix_set}).
Such products have been studied for several decades~\cite{furstenberg1960products} 
and have found applications in physics, biology and economics. Here they provide 
a convenient way to explicitly model the stochasticity of the swarm dynamics such that
we can claim that the performance of PSO is determined by the stability properties
of the random dynamical system (\ref{eq:pso_expl-1}). 

\begin{figure}[th]
\begin{center}
\includegraphics[width=0.6\linewidth]{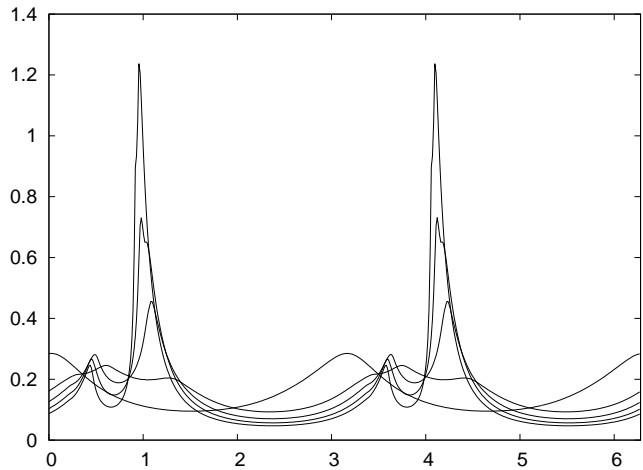}
\end{center}
\vspace{-5mm}
\caption{Stationary distribution $\nu_{\alpha,\omega}(\mathbf{a})$ on the unit circle 
	($\mathbf{a} \in [0,2\pi)$)
in the $\left(x,v\right)$ plane for a one-particle system (\ref{eq:pso_expl-1}) for
$\omega=0.7$ and $\alpha=\alpha_2=0.5$, $1.5$, $2.5$, $3.5$, $4.5$ (the
distribution with peak near $\pi$ is for $\alpha=0.5$, otherwise
main peaks are highest for largest $\alpha$). 
\label{fig:Stationary-distribution}}
\end{figure}

\begin{figure}[th]
\begin{center}
\includegraphics[width=0.65\linewidth]{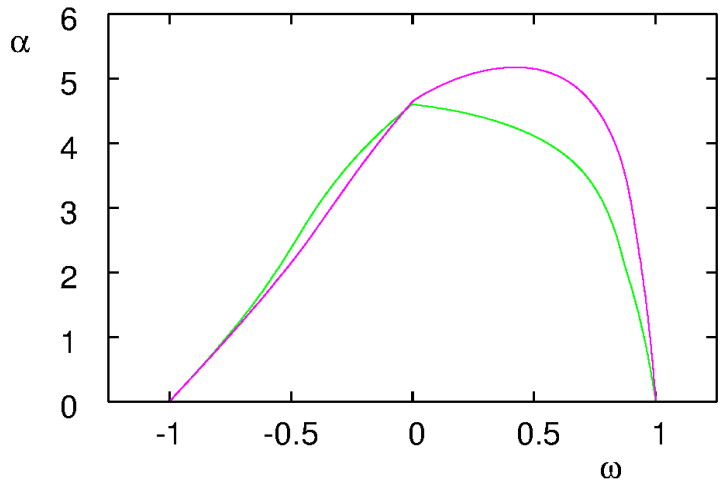}
\vspace{-4mm}
\end{center}
\vspace{-5mm}
\caption{Solution of Eq.~\ref{eq:implicite_a_w} representing 
	a single particle in one dimension with a fixed best value at $\mathbf{g}=\mathbf{p}=0$.
	The curve that has higher $\alpha$-values on the right (magenta) is for $\alpha_1=\alpha_2$,
	the other curve (green) is for $\alpha=\alpha_2$, $\alpha_1=0$.
Except for the regions near $\omega =\pm 1$, where numerical instabilities can occur, 
a simulation produces an indistinguishable curve. In the simulation we tracked the probability of 
a particle to either reach a small region 
($10^{-6}$) near the origin or to escape beyond a radius of $10^6$ after starting from a random location
on the unit circle. Along the curve both probabilities are equal.
\label{fig:Theory_and_Numerical}}
\end{figure}

\begin{figure}[th]
\begin{center}
\includegraphics[width=0.65\linewidth]{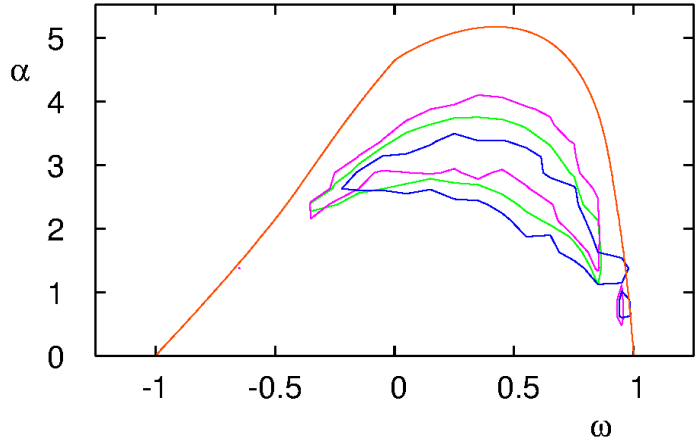}
\end{center}
\vspace{-5mm}
\caption{
Best parameter regions for 200 (blue), 2000 (green), and 20000 (magenta) iterations: For more
iterations the region shifts towards the critical line.
Cost averaged over 100 runs and 28 CEC benchmark functions. 
The red (outer) curve represents the zero Lyapunov exponent for $N=1$, $d=1$, $\alpha_{1}=\alpha_{2}$.
\label{fig:PSOAverageValley}}
\end{figure}

Since the equation (\ref{eq:pso_expl-1}) is linear,
the analysis can be restricted to vectors on the unit sphere in the  
$(\mathbf{v},\mathbf{x})$ space, i.e.~to unit vectors
\begin{equation}
	\mathbf{a}=\left(\mathbf{x},\mathbf{v}\right)^{\top}\!/\,
	\Vert \left(\mathbf{x},\mathbf{v}\right)^{\top}\!\Vert,
\label{eq:unit_circle}
\end{equation}
where $\Vert \cdot \Vert $ denotes the Euclidean norm. Unless the set of matrices shares the same 
eigenvectors (which is not the case here) standard stability analysis in terms of
eigenvalues is not applicable. Instead we will use means from the theory of 
random matrix products in order to decide
whether the set of matrices is stochastically contractive.
The properties of the asymptotic dynamics can be described based on a double
Lebesgue integral over the unit sphere $S^{2d-1}$ and the set 
$\cal{M}$~\cite{tutubalin1965limit,khas1967necessary}. As in Lyapunov exponents, 
the effect of the dynamics
is measured in logarithmic units in order to account for multiplicative action.
\begin{equation}
	\lambda\left(\alpha,\omega\right)=\int\! d\nu_{\alpha,\omega}\left(\mathbf{a}\right)\!\int dP_{\alpha,\omega}\left(M\right)\,\log\left\Vert M\mathbf{a}\right\Vert \label{eq:lyapunov}
\end{equation}
If $a\left(\alpha,\omega\right)$ is negative the algorithm will converge to
$\mathbf{p}$ with probability 1, while for positive $a$ arbitrarily large fluctuations are possible.
While the measure for the inner integral (\ref{eq:lyapunov}) is given by Eq.~\ref{eq:convolution},
we have to determine the stationary distribution $\nu$ on the unit sphere for 
the outer integral. It is given as the solution of the integral equation
\begin{equation}
	\nu_{\alpha,\omega}\left(\mathbf{a}\right)=\int\! d\nu_{\alpha,\omega}\left(\mathbf{b}\right)\int\! dP_{\alpha,\omega}\left(M\right)\delta\left(\mathbf{a},M\mathbf{b} / \left\Vert M\mathbf{b}\right\Vert \right),\,\,\,\mathbf{a},\mathbf{b}\in S^{2d-1}.\label{eq:stationary_measure}
\end{equation}
The existence of the invariant measure requires the dynamics to be ergodic which is ensured if
at least some of elements of $\cal{M}$ have complex eigenvalues, such as being the case for
$\omega^2+\alpha^2/4-\omega\alpha-2\omega-\alpha+1<0$ (see above, \cite{trelea2003particle}). This condition
excludes a small region in the parameters space at small values of $\omega$, such that there we have to
take all ergodic components into account. There are not more than two components which due to symmetry 
have the same stability properties.
It depends on the parameters $\alpha$ and $\omega$ and differs strongly from a homogenous distribution, see Fig.~\ref{fig:Stationary-distribution} for a few examples in the case $d=1$. 
Critical parameters are obtained from Eq.~\ref{eq:lyapunov} by the relation
\begin{equation}
\lambda\left(\alpha,\omega\right)=0\label{eq:implicite_a_w}.
\end{equation}
Solving Eq.~\ref{eq:implicite_a_w} is difficult in higher dimensions, so we rely on the 
linearity of the system when considering the $(d=1)$-case as representative.
The curve in Fig.~\ref{fig:Theory_and_Numerical} represents the solution of Eq.~\ref{eq:implicite_a_w}
for $d=1$ and $\alpha=\alpha_2$. For other settings of $\alpha_1$ and $\alpha_2$ the distribution
of the random factors has a smaller variance rendering the dynamics more stable such that the contour
moves towards larger parameter values (see Fig.~\ref{fig:PSOAverageValley}).
Inside the contour $\lambda\left(\alpha,\omega\right)$ is negative, meaning 
that the state will approach the origin with probability 1. Along the contour and in the outside
region large state fluctuations are possible. Interesting parameter values are expected near the
curve where due to a coexistence of stable and unstable dynamics (induced by different sequences
of random matrices) a theoretically 
optimal combination of exploration and exploitation is possible. For specific  
problems, however, deviations from the critical curve can be expected to be beneficial.

\subsection{Personal best vs.~global best}
\label{Personalvsglobal}

\begin{figure}[th]
\begin{center}
\includegraphics[width=0.65\linewidth,trim=0cm 3.5mm 0cm 0cm]{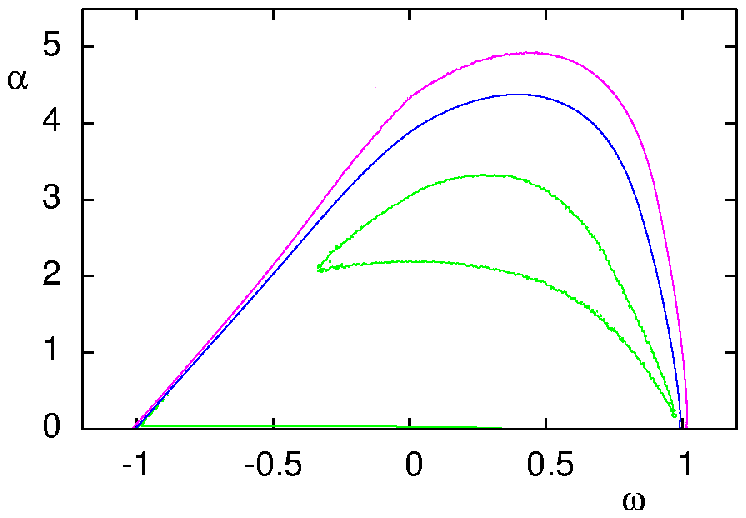}
\end{center}
\vspace{-5mm}
\caption{
For $\mathbf{p}\ne\mathbf{g}$ we define neutral stability as the equilibrium between
divergence and convergence. Convergence means here that the particle approaches the line
connecting $\mathbf{p}$ and $\mathbf{g}$.
Curves are for a one-dimensional problem with $\mathbf{p}=0.1$ and $\mathbf{g}=0$ scaled (see 
Sect.~\ref{Personalvsglobal}) by $\kappa=1$ (outer curve) $\kappa=0.1$ and $\kappa=0.04$ (inner curve).
Results are for 200 iterations and averaged over 100000 repetitions. 
\label{fig:PSOAverageValley_p}}
\end{figure}

Due to linearity, the particle swarm update rule (\ref{eq:PSOPositionUpdate}) is subject
to a scaling invariance which was already used in Eq.~\ref{eq:unit_circle}. 
We now consider the consequences of linearity for the case where 
personal best and global best differ, i.e.~$\mathbf{p}\ne \mathbf{g}$. 
For an interval where $\mathbf{p}_i$ and $\mathbf{g}$ remain unchanged, 
the particle $i$ with personal best $\mathbf p_i$ will behave like a particle in a
swarm where together with $\mathbf{x}$ and $\mathbf{v}$,  $\mathbf{p}_i$ is also
scaled by a factor $\kappa>0$. The finite-time approximation of the
Lyapunov exponent (see Eq.~\ref{eq:lyapunov})
\begin{equation}
	\lambda(t)=\frac{1}{t} \log \left\langle\left\| \left(\mathbf{x}_t,\mathbf{v}_t\right)\right\| \right\rangle
\end{equation}
will be changed by an amount of $\frac{1}{t} \log \kappa $ by the scaling.
Although this has no effect on the asymptotic behaviour, we will have to expect 
an effect on the stability of the swarm for finite times which may be relevant for
practical applications. For the same parameters, the swarm will be more stable
if $\kappa<1$ and less stable for $\kappa>1$, 
provided that the initial conditions are scaled in the same way. 
Likewise, if $\|\mathbf{p}\|$ is increased,
then the critical contour will move inwards, see Fig.~\ref{fig:PSOAverageValley_p}.
Note that in this figure, the low number of iterations lead
to a few erroneous trials at parameter pairs outside the outer contour which have been omitted here.
We also do not consider the behaviour near $\alpha=0$ which is complex but irrelevant for PSO.
The contour (\ref{eq:implicite_a_w}) can be seen as the limit $\kappa \to 0$ such that
only an increase of $\|\mathbf{p}\|$ is relevant for comparison with the theoretical
stability result. When comparing the stability results with numerical simulations for
real optimisation problems, we will need to take into account the effects caused by differences between $\mathbf{p}$ and $\mathbf{g}$ in a multi-particle swarm with finite runtimes.

\section{Optimisation of benchmark functions\label{Simulations}}

Metaheuristic algorithms are often tested in competition against benchmark functions designed to 
present different problem space characteristics. 
The 28 functions~\cite{CEC2013} contain a mix of unimodal, basic multimodal and composite functions. 
The domain of the functions in this test set are all defined to be $[-100, 100]^d$ 
where $d$ is the dimensionality of the problem. Particles were initialised within the same domain.
We use 10-dimensional problems throughout. 
Our implementation of PSO performed no spatial or velocity clamping.  In all trials a swarm of 25 particles was used. 
We repeated the algorithm 100 times, on each occasion allowing 200, 2000, 20000 iterations to pass before 
recording the best solution found by the swarm. For the competition 50000 fitness evaluation were allowed which
corresponds to 2000 iterations with 25 particles. Other iteration numbers were included for comparison.
This protocol was carried out for pairs of $\omega\in[-1.1,1.1]$ and $\alpha\in[0,5]$ 
This was repeated for all 28 functions. 
The averaged solution costs as a function of the two parameters 
showed curved valleys similar to that in Fig.~\ref{fig:PSOPerformance} for all problems. 
For each function we obtain different best values along (or near) the theoretical curve 
(\ref{eq:implicite_a_w}). There appears to be no
preferable location within the valley. 
Some individual functions yield best performance 
near $\omega=1$. This is not the case near $\omega=0$, although the global average performance 
over all test functions is better in the valley near  $\omega=0$ than near $\omega=1$, see
Fig~\ref{fig:PSOAverageValley}.

At medium values of $\omega$ 
the difference between the analytical solutions for the cases 
$\alpha_1=\alpha_2$ and $\alpha_1=0$ is strongest, see Fig.~\ref{fig:PSOAverageValley}. 
In simulations this shows to a lesser extent, thus revealing a shortcoming of the one-particle
approximation. Because in the multi-particle case, $\mathbf{p}$ and $\mathbf{g}$ are often
different, the resulting vector will have a smaller norm than in the one-particle case, where 
$\mathbf{p}=\mathbf{g}$. The case $\mathbf{p}\ne\mathbf{g}$ violates a the assumption of
the theory the dynamics can be described based unit vectors. While a particle far away from
both $\mathbf{p}$ and $\mathbf{g}$ will behave as predicted from the one-particle case,
at length scales smaller than $\Vert \mathbf{p}-\mathbf{g}\Vert$ the retractive forces will
tend to be reduced such that the inertia becomes more effective and the particle is locally less
stable which shows numerically in optimal parameters that are smaller than predicted.

\section{Discussion\label{discussion}}

\subsection{Relevance of criticality}

Our analytical approach predicts a locus of $\alpha$ and $\omega$ pairings that maintain the critical
behaviour of the PSO swarm.
Outside this line the swarm will diverge unless steps are taken to constrain it. Inside, the swarm 
will eventually converge to a single solution.
In order to locate a solution within the search space, the swarm needs to converge at some point, so the line represents an upper bound on the exploration-exploitation mix that a swarm manifests. 
For parameters on the critical line, fluctuations are still arbitrary large. Therefore, subcritical 
parameter values can be preferable if the settling time is of the same order as the scheduled 
runtime of the algorithm. If, in addition, a typical length scale of the problem is known, then the
finite standard deviation of the particles in the stable parameter region 
can be used to decide about the distance of the 
parameter values from the critical curve. These dynamical quantities can be approximately set, 
based on the theory presented here, such that a precise control of the behaviour of the algorithm is in
principle possible. 

The observation of the distribution of empirically optimal parameter values along the critical curve,
confirms the expectation that critical or near-critical behaviour is the main reason for success
of the algorithm. Critical fluctuations are a plausible tool in search problem if apart from
certain smoothness assumption nothing is known about the cost landscape: The majority of excursions
will exploit the smoothness of the cost function by local search, whereas the fat tails of the 
distribution allow the particles to escape from local minima.

\subsection{Switching dynamics at discovery of better solutions\label{sub:The-role-of}}

Eq.~\ref{eq:matrix_from} shows that the discovery of a better solution
affects only the constant terms of the linear dynamics of a particle, whereas its
dynamical properties are governed by the linear coefficient matrices. 
However, in the time step after a particle has found a new solution the corresponding force term 
in the dynamics is zero (see Eq.~\ref{eq:PSOPositionUpdate}) such that the particle dynamics 
slows down compared to the theoretical solution which assumes a finite
distance from the best position at all (finite) times. As this affects usually only one particle
at a time and because new discoveries tend to become rarer over time, this effect will be small
in the asymptotic dynamics, although it could justify the empirical optimality of parameters 
in the unstable region for some test cases.

The question is nevertheless, how often these changes occur. A weakly
converging swarm can still produce good results if it often discovers
better solutions by means of the fluctuations it performs before settling
into the current best position. For cost functions that are not `deceptive',
i.e.~where local optima tend to be near better optima, parameter
values far inside the critical contour (see Fig.~\ref{fig:Theory_and_Numerical})
may give good results, while in other cases more exploration is needed.

\subsection{The role of personal best and global best\label{current_best}\label{runtime}}

A numerical scan of the $(\alpha_1,\alpha_2)$ plane shows 
a valley of good fitness values, which, at small fixed positive $\omega$, is roughly linear and described 
by the relation $\alpha_1+\alpha_2= \mbox{const}$, i.e.~only the joint parameter
$\alpha=\alpha_1+\alpha_2$ matters.
For large $\omega$, and accordingly small predicted optimal $\alpha$ values, 
the valley is less straight. This may be 
because the effect of the known solutions is 
relatively weak, so the interaction of the two components becomes more important.
In other words if the movement of the particles is mainly due to inertia, 
then the relation between the global and local best is non-trivial, 
while at low inertia the particles can adjust their $\mathbf{p}$ vectors
quickly towards the  $\mathbf{g}$ vector such that both terms become interchangeable.

Finally, we should mention that more particles, longer runtime as well as lower search 
space dimension increase the potential for 
exploration. They all lead to the empirically determined optimal parameters being closer 
to the critical curve.

\section{Conclusion}

PSO is a widely used optimisation scheme which is theoretically not well understood. Existing theory 
concentrates on a deterministic version of the algorithm which does not possess useful exploration
capabilities. We have studied the algorithm by means of a product of random matrices which allows us to
predict useful parameter ranges and may allow for more precise settings 
if a typical length scale of the problem is known.
A weakness of the current approach is that it focuses on the standard PSO~\cite{kennedy1995particle} which
is known to include biases~\cite{clerc2006confinments,spears2010biases}, 
that are not necessarily justifiable,
and to be outperformed on benchmark set and in practical applications by many of the existing PSO variants.
Similar analyses are certainly possible and are expected to be carried out for some of the variants, 
even though the field of metaheuristic search is often portrayed as largely inert to theoretical advances.
If the dynamics of particle swarms is better understood, the algorithms may become useful as
efficient particle filters which have many applications beyond heuristic optimisation.

\subsection*{Acknowledgments}

This work was supported by the Engineering and Physical Sciences Research Council (EPSRC), grant number EP/K503034/1.

\bibliographystyle{unsrt}
\bibliography{pso_JMH}

\end{document}